\documentclass[runningheads]{llncs}
\usepackage{times}
\usepackage{epsfig}
\usepackage{graphicx}
\usepackage{amsmath}
\usepackage{amssymb}
\usepackage{commath}
\usepackage{multirow}
\usepackage{algorithmic}
\usepackage[norelsize,longend,ruled,lined,boxed,commentsnumbered]{algorithm2e}
\usepackage{breqn}
\usepackage{color}
\usepackage{caption}
\usepackage{subcaption}

\usepackage{fancyhdr}
\fancypagestyle{specialfooter}{%
  \fancyhf{}
  
  \fancyhead[R]{\scriptsize{Accepted in CVMI 2022 Conference}}
  \fancyfoot[L]{\scriptsize{International Conference on Computer Vision and Machine Intelligence (CVMI), 2022}}
}

\begin{document}
\title{Moment Centralization based Gradient Descent Optimizers for Convolutional Neural Networks}

\author{Sumanth Sadu\inst{1} \and Shiv Ram Dubey\inst{2} \and S. R. Sreeja \inst{3}}
\institute{Computer Vision Group, Department of Computer Science and Engineering, Indian Institute of Information Technology, Sri City, Andhra Pradesh, India
\and
Computer Vision and Biometrics Laboratory, Indian Institute of Information Technology, Allahabad, Uttar Pradesh, India\\
\and
Department of Computer Science and Engineering, Indian Institute of Information Technology, Sri City, Andhra Pradesh, India \\
{\tt\small venkatasaisumanth.s18@iiits.in, srdubey@iiita.ac.in, sreeja.sr@iiits.in}
}

\maketitle
\thispagestyle{specialfooter}

\begin{abstract}
Convolutional neural networks (CNNs) have shown very appealing performance for many computer vision applications. The training of CNNs is generally performed using stochastic gradient descent (SGD) based optimization techniques. The adaptive momentum-based SGD optimizers are the recent trends. However, the existing optimizers are not able to maintain a zero mean in the first-order moment and struggle with optimization. In this paper, we propose a moment centralization-based SGD optimizer for CNNs. Specifically, we impose the zero mean constraints on the first-order moment explicitly. The proposed moment centralization is generic in nature and can be integrated with any of the existing adaptive momentum-based optimizers. The proposed idea is tested with three state-of-the-art optimization techniques, including Adam, Radam, and Adabelief on benchmark CIFAR10, CIFAR100, and TinyImageNet datasets for image classification. The performance of the existing optimizers is generally improved when integrated with the proposed moment centralization. Further, The results of the proposed moment centralization are also better than the existing gradient centralization. The analytical analysis using the toy example shows that the proposed method leads to a shorter and smoother optimization trajectory. The source code is made publicly
available at \url{https://github.com/sumanthsadhu/MC-optimizer}.
\end{abstract}

\section{Introduction}
Deep learning has become very prominent to solve the problems in computer vision, natural language processing, and speech processing \cite{deeplearning}. Convolutional neural networks (CNNs) have been exploited to utilize deep learning in the computer vision area with great success to deal with the image and video data \cite{alexnet}, \cite{vgg}, \cite{resnet}, \cite{srivastava2020hard}, \cite{basha2018rccnet}, \cite{repala2019dual}. The CNN models are generally trained using the stochastic gradient descent (SGD) \cite{sgd} based optimization techniques where the parameters of the model are updated in the opposite direction of the gradient. The performance of any CNN model is also sensitive to the chosen SGD optimizer used for the training. Hence, several SGD optimizers have been investigated in the literature.

The vanilla SGD algorithm suffers from the problem of zero gradients at local minimum and saddle regions. The SGD with momentum (SGDM) uses the accumulated gradient, i.e., momentum, to update the parameters instead of the current gradient \cite{sgdm}. Hence, it is able to update the parameters in the local minimum and saddle regions. However, the same step size is used by SGDM for all the parameters. In order to adapt the step size based on the gradient consistency, AdaGrad \cite{adagrad} divides the learning rate by the square root of the accumulated squared gradients from the initial iteration. However, it leads to a vanishing learning rate after some iteration as the squared gradient is always positive. In order to tackle the learning rate diminishing problem of AdaGrad, RMSProp \cite{rmsprop} uses a decay factor on the accumulated squared gradient. The idea of both SGDM and RMSProp is combined in a very popular Adam gradient descent optimizer \cite{adam}. Basically, Adam uses a first order moment as the accumulated gradient for parameter update and a second order moment as the accumulated squared gradient to control the learning rate. A rectified adam (Radam) \cite{radam} performs the rectification in Adam to switch to SGDM to improve the precision of the convergence of training. Recently, Adabelief optimizer \cite{adabelief} utilizes the residual of gradient and first order moment to compute the second order moment which improves the training at saddle regions and local minimum. Other notable recent optimizers include diffGrad \cite{diffgrad}, AngularGrad \cite{angulargrad}, signSGD \cite{signsgd}, Nostalgic Adam \cite{nostalgicadam}, and AdaInject \cite{adainject}.

The above mentioned optimization techniques do not utilize the normalization of the moment for smoother training. However, it is evident that normalization plays a very important role in the training of deep learning models, such as data normalization \cite{singh2020investigating}, batch normalization \cite{batchnormalization}, instance normalization \cite{choi2021meta}, weight normalization \cite{salimans2016weight}, and gradient normalization \cite{chen2018gradnorm}.
The gradient normalization is also utilized with SGD optimizer in \cite{gc}. However, as the first order moment is used for the parameter updates in adaptive momentum based optimizers, we proposed to perform the normalization on the accumulated first order moment. The contributions of this paper can be summarized as follows:
\begin{itemize}
    \item We propose the concept of moment centralization to normalize the first order moment for the smoother training of CNNs.
    \item The proposed moment centralization is integrated with Adam, Radam, and Adabelief optimizers.
    \item The performance of the existing optimizers is tested with and without the moment centralization.
\end{itemize}

The rest of the paper is structured as follows: Section 2 presents the proposed moment centralization based SGD optimizers; Section 3 summarizes the experimental setup, datasets used and CNN models used; Section 4 illustrates the experimental results, comparison and analysis; and finally Section 5 concludes the paper.

\section{Proposed Moment Centralization based SGD Optimizers}
In this paper, a moment centralization strategy is proposed for the adaptive momentum based SGD optimizers for CNNs. The proposed approach imposes the explicit normalization over the first order moment in each iteration. The imposed zero mean distribution helps the CNN models for a smoother training. The proposed moment centralization is generic and can be integrated with any adaptive momentum based SGD optimizer. In this paper, we use it with state-of-the-art optimizers, including Adam \cite{adam}, Radam \cite{radam} and Adabelief \cite{adabelief}. The Adam optimizer with moment centralization is termed AdamMC. Similarly, the Radam and Adabelief optimizers with moment centralization are referred to as RadamMC and AdabeliefMC, respectively. In this section, we explain the proposed concept with Adam, i.e., AdamMC.

Let a function $f$ represents a CNN model having parameters $\theta$ for image classification. Initially, the parameters $\theta_0$ are initialized randomly and updated using backpropagation of gradients in the subsequent iterations of training.
During the forward pass of training, the CNN model takes a batch of images ($I_B$) containing $B$ images as the input and returns the cross-entropy loss as follows:
\begin{equation}
    L_{CE} = -\sum_{i=1}^{B} \log(p_{c_i})
\end{equation}
where $p_{c_i}$ is the softmax probability of given input image $I_i$ for the correct class level $c_i$. Note that the class level is represented by the one-hot encoding in the implementation. During the backward pass of the training, the parameters are updated based on the gradient information. Different SGD optimizers utilize the gradient information differently for the parameter update. The adaptive momentum based optimizers such as Adam are very common in practice which utilizes the first order and second order momentum which is the exponential moving average of gradient and squared gradient, respectively. Hence, we explain the proposed concept below with the help of Adam optimizer.

Consider $g_t$ to be the gradient of an objective function w.r.t. the parameter $\theta$, i.e., $g_t = \nabla_{\theta} f_t(\theta_{t-1})$ in the $t^{th}$ iteration of training, where $\theta_{t-1}$ represents the parameter values obtained after $(t-1)^{th}$ iteration. The first order moment ($m_t$) in the $t^{th}$ iteration is computed as follows,
\begin{equation}
    m_t = \beta_1 m_{t-1} + (1-\beta_1) g_t
\end{equation}
where $m_{t-1}$ is the first order moment in $(t-1)^{th}$ iteration and $\beta_1$ is a decay hyperparameter. Note that the distribution of first order moments is generally not zero centric which leads to the update in a similar direction for the majority of the parameters. In order to tackle this problem, we propose to perform the moment centralization by normalizing the first order moment with zero mean as follows,
\begin{equation}
    m_t = m_t  - mean(m_t).
\end{equation}
The second order moment ($v_t$) in the $t^{th}$ iteration is computed as follows,
\begin{equation}
    v_t \gets \beta_2 v_{t-1} + (1-\beta_2) g^2_t
\end{equation}
where $v_{t-1}$ is the second order moment in $(t-1)^{th}$ iteration and $\beta_2$ is a decay hyperparameter. Note that we do not perform the centralization of second order moments as it may lead to a very inconsistent effective learning rate.

\begin{algorithm}[!t]
\caption{Adam Optimizer \cite{adam}}
\SetAlgoLined
\textbf{Initialize:} $\theta_{0},m_{0}\gets0,v_{0}\gets0,t\gets0$ \hspace{10mm} \textbf{Hyperparameters:} $\alpha, \beta_1, \beta_2, T$\\
\textbf{For} $\theta_{t}=$ [1,...,\textit{T}] \textbf{do}\\
    \hspace{0.45cm} $g_t \gets \nabla_{\theta} f_t(\theta_{t-1})$   \hspace{3.85cm} \textit{$\rhd$ Gradient Computation}\\
    \hspace{0.45cm} $m_t \gets \beta_1 m_{t-1} + (1-\beta_1) g_t$  \hspace{2.4cm} \textit{$\rhd$ First Order Moment}\\
    \hspace{0.45cm} $v_t \gets \beta_2 v_{t-1} + (1-\beta_2) g^2_t$  \hspace{2.62cm} \textit{$\rhd$ Second Order Moment}\\
    \hspace{0.45cm} $\widehat{m_t} \gets m_t/(1-\beta_1^t)$, \hspace{3mm} $\widehat{v_t} \gets v_t/(1-\beta_2^t)$ \hspace{0.68cm} \textit{$\rhd$ Bias Correction}\\
    \hspace{0.5cm} $\theta_t \gets \theta_{t-1} - \alpha \widehat{m_t}/(\sqrt{\widehat{v_t}} + \epsilon)$ \hspace{2.18cm} \textit{ 	$\rhd$ Parameter Update}\\
    \textbf{return}  $\theta_{T}$
\label{algo:adam}
\end{algorithm}

\begin{algorithm}[!t]
\caption{Adam + Moment Centralization (AdamMC) Optimizer}
\SetAlgoLined
\textbf{Initialize:} $\theta_{0},m_{0}\gets0,v_{0}\gets0,t\gets0$ \hspace{10mm} \textbf{Hyperparameters:} $\alpha, \beta_1, \beta_2, T$\\
\textbf{For} $\theta_{t}=$ [1,...,\textit{T}] \textbf{do}\\
    \hspace{0.45cm} $g_t \gets \nabla_{\theta} f_t(\theta_{t-1})$   \hspace{3.85cm} \textit{$\rhd$ Gradient Computation}\\
    \hspace{0.45cm} $m_t \gets \beta_1 m_{t-1} + (1-\beta_1) g_t$  \hspace{2.4cm} \textit{$\rhd$ First Order Moment}\\
    \hspace{0.45cm} \textcolor{blue} {$m_t \gets m_t  - mean(m_t)$}  \hspace{3.01cm} \textit{$\rhd$ \textcolor{blue}{Moment Centralization}}\\
    \hspace{0.45cm} $v_t \gets \beta_2 v_{t-1} + (1-\beta_2) g^2_t$  \hspace{2.62cm} \textit{$\rhd$ Second Order Moment}\\
    \hspace{0.45cm} $\widehat{m_t} \gets m_t/(1-\beta_1^t)$, \hspace{3mm} $\widehat{v_t} \gets v_t/(1-\beta_2^t)$ \hspace{0.68cm} \textit{$\rhd$ Bias Correction}\\
    \hspace{0.5cm} $\theta_t \gets \theta_{t-1} - \alpha \widehat{m_t}/(\sqrt{\widehat{v_t}} + \epsilon)$ \hspace{2.18cm} \textit{ 	$\rhd$ Parameter Update}\\
    \textbf{return}  $\theta_{T}$
\label{algo:adammc}
\end{algorithm}

The first and second order moments suffer in the initial iteration due to very small values leading to very high effective step size. In order to cope up with this bias, Adam \cite{adam} has introduced a bias correction as follows,
\begin{equation}
    \widehat{m_t} \gets m_t/(1-\beta_1^t), \hspace{3mm} \widehat{v_t} \gets v_t/(1-\beta_2^t)
\end{equation}
where $\widehat{m_t}$ and $\widehat{v_t}$ are the bias corrected first and second order moment, respectively. Finally, the parameter update in $t^{th}$ iteration is performed as follows,
\begin{equation}
    \theta_t \gets \theta_{t-1} - \alpha \widehat{m_t}/(\sqrt{\widehat{v_t}} + \epsilon)
\end{equation}
where $\theta_t$ is the updated parameter, $\theta_{t-1}$ is the parameter after $(t-1)^{th}$ iteration, $\alpha$ is the learning rate and $\epsilon$ is a very small constant number used for numerical stability to avoid the division by zero.
The steps of Adam optimizer without and with the proposed moment centralization concept are summarized in Algorithm \ref{algo:adam} (Adam) and Algorithm \ref{algo:adammc} (AdamMC), respectively. The changes in the AdamMC are highlighted in blue color. Similarly, we also incorporate the moment centralization concept with Radam \cite{radam} and Adabelief \cite{adabelief} optimizers. The steps for Radam, RadamMC, Adabelief, and AdabeliefMC are illustrated in Algorithm \ref{algo:radam}, \ref{algo:radammc}, \ref{algo:adabelief}, and \ref{algo:adabeliefmc}, respectively.

\begin{algorithm}[!t]
\caption{Radam Optimizer \cite{radam}}
\SetAlgoLined
\textbf{Initialize:} $\theta_{0},m_{0}\gets0,v_{0}\gets0$ \hspace{10mm} \textbf{Hyperparameters:} $\alpha, \beta_1, \beta_2, T$\\
\textbf{For} $\theta_{t}=$ [1,...,\textit{T}] \textbf{do}\\
    \hspace{0.45cm} $g_t \gets \nabla_{\theta} f_t(\theta_{t-1})$      \hspace{2.75cm} \textit{$\rhd$ Gradient Computation} \\
    \hspace{0.45cm} $m_t \gets \beta_1 m_{t-1} + (1-\beta_1) g_t$ \hspace{1.28cm} \textit{$\rhd$ First Order Moment} \\
    \hspace{0.45cm} $v_t \gets \beta_2 v_{t-1} + (1-\beta_2) g_t^2$ \hspace{1.5cm} \textit{$\rhd$ Second Order Moment} \\
    \hspace{0.45cm} $\rho_\infty \gets 2 / (1 - \beta_2) - 1$ \\
    \hspace{0.45cm} $\rho_t = \rho_\infty - 2t\beta_2^t/(1-\beta_2^t)$\\
    \hspace{0.45cm} \textbf{If} $\rho_t \geq 5$ \hspace{3.88cm} \textit{$\rhd$ Check if the variance is tractable}\\
    \hspace{0.9cm} $\rho_u = (\rho_t - 4) \times (\rho_t - 2) \times \rho_\infty$\\
    \hspace{0.9cm} $\rho_d = (\rho_\infty - 4) \times (\rho_\infty - 2) \times \rho_t$\\
    \hspace{0.9cm} $\rho = \sqrt{(1 - \beta_2) \times \rho_u / \rho_d}$ \hspace{1.32cm} \textit{$\rhd$ Variance rectification term}\\
    \hspace{0.9cm} $\alpha_1 = \rho \times \alpha / (1 - \beta_1^t)$ \hspace{1.75cm} \textit{$\rhd$ Rectified learning rate}\\
    \hspace{0.9cm} $\theta_t \gets \theta_{t-1} - \alpha_1 m_t/(\sqrt{v_t} + \epsilon)$ \hspace{0.6cm} \textit{$\rhd$ Update parameters with rectification}\\
    \hspace{0.45cm} \textbf{Else}\\
    \hspace{0.9cm} $\alpha_2 = \alpha / (1 - \beta_1^t)$ \hspace{2.3cm} \textit{$\rhd$ Bias correction}\\
    \hspace{0.9cm} $\theta_t \gets \theta_{t-1} - \alpha_2 m_t$  \hspace{2.06cm} \textit{$\rhd$ Update parameters without rectification}
    \textbf{return}  $\theta_{T}$
\label{algo:radam}
\end{algorithm}

\begin{algorithm}[!t]
\caption{Radam + Moment Centralization (RadamMC) Optimizer}
\SetAlgoLined
\textbf{Initialize:} $\theta_{0},m_{0}\gets0,v_{0}\gets0$ \hspace{10mm} \textbf{Hyperparameters:} $\alpha, \beta_1, \beta_2, T$\\
\textbf{For} $\theta_{t}=$ [1,...,\textit{T}] \textbf{do}\\
    \hspace{0.45cm} $g_t \gets \nabla_{\theta} f_t(\theta_{t-1})$      \hspace{2.75cm} \textit{$\rhd$ Gradient Computation} \\
    \hspace{0.45cm} $m_t \gets \beta_1 m_{t-1} + (1-\beta_1) g_t$ \hspace{1.28cm} \textit{$\rhd$ First Order Moment} \\
    \hspace{0.45cm} \textcolor{blue} {$m_t \gets m_t  - mean(m_t)$}  \hspace{1.9cm} \textit{$\rhd$  \textcolor{blue}{Moment Centralization}}\\
    \hspace{0.45cm} $v_t \gets \beta_2 v_{t-1} + (1-\beta_2) g_t^2$ \hspace{1.5cm} \textit{$\rhd$ Second Order Moment} \\
    \hspace{0.45cm} $\rho_\infty \gets 2 / (1 - \beta_2) - 1$ \\
    \hspace{0.45cm} $\rho_t = \rho_\infty - 2t\beta_2^t/(1-\beta_2^t)$\\
    \hspace{0.45cm} \textbf{If} $\rho_t \geq 5$ \hspace{3.88cm} \textit{$\rhd$ Check if the variance is tractable}\\
    \hspace{0.9cm} $\rho_u = (\rho_t - 4) \times (\rho_t - 2) \times \rho_\infty$\\
    \hspace{0.9cm} $\rho_d = (\rho_\infty - 4) \times (\rho_\infty - 2) \times \rho_t$\\
    \hspace{0.9cm} $\rho = \sqrt{(1 - \beta_2) \times \rho_u / \rho_d}$ \hspace{1.32cm} \textit{$\rhd$ Variance rectification term}\\
    \hspace{0.9cm} $\alpha_1 = \rho \times \alpha / (1 - \beta_1^t)$ \hspace{1.75cm} \textit{$\rhd$ Rectified learning rate}\\
    \hspace{0.9cm} $\theta_t \gets \theta_{t-1} - \alpha_1 m_t/(\sqrt{v_t} + \epsilon)$ \hspace{0.6cm} \textit{$\rhd$ Update parameters with rectification}\\
    \hspace{0.45cm} \textbf{Else}\\
    \hspace{0.9cm} $\alpha_2 = \alpha / (1 - \beta_1^t)$ \hspace{2.3cm} \textit{$\rhd$ Bias correction}\\
    \hspace{0.9cm} $\theta_t \gets \theta_{t-1} - \alpha_2 m_t$  \hspace{2.06cm} \textit{$\rhd$ Update parameters without rectification}
    \textbf{return}  $\theta_{T}$
\label{algo:radammc}
\end{algorithm}

\section{Experimental Setup}
\label{experimental_setup}
All experiments are conducted on Google Colab GPUs using the Pytorch 1.9 framework. We want to emphasise that our Moment Centralization (MC) approach does not include any additional hyper-parameters. To incorporate MC into the existing optimizers, only one line of code is required to be included in the code of existing adaptive momentum based SGD optimizers, with all other settings remaining untouched.

\subsection{CNN Models Used}
We use VGG16 \cite{vgg} and ResNet18 \cite{resnet} CNN models in the experiments to validate the performance of the proposed moment centralization based optimizers, including AdamMC, RadamMC and AdabeliefMC. The VGG16 is a plain CNN model with sixteen learnable layers. It uses three fully connected layers towards the end of the network. It is one of the popular CNN models utilized for different computer vision tasks. The ResNet18 is a directed acyclic graph-based CNN model which utilizes the identity or residual connections in the network. The identity connection improves the gradient flow in the network during backpropagation and helps in the training of a deep CNN model.

\begin{algorithm}[!t]
\caption{Adabelief Optimizer \cite{adabelief}}
\SetAlgoLined
\textbf{Initialize:} $\theta_{0},m_{0}\gets0,v_{0}\gets0,t\gets0$ \hspace{10mm} \textbf{Hyperparameters:} $\alpha, \beta_1, \beta_2, T$\\
\textbf{For} $\theta_{t}=$ [1,...,\textit{T}] \textbf{do}\\
    \hspace{0.45cm} $g_t \gets \nabla_{\theta} f_t(\theta_{t-1})$   \hspace{3.85cm} \textit{$\rhd$ Gradient Computation}\\
    \hspace{0.45cm} $m_t \gets \beta_1 m_{t-1} + (1-\beta_1) g_t$  \hspace{2.40cm} \textit{$\rhd$ First Order Moment}\\
    \hspace{0.45cm} $v_t \gets \beta_2 v_{t-1} + (1-\beta_2) (g_t-m_t)^2$ \hspace{1.45cm} \textit{$\rhd$ Second Order Moment} \\
    \hspace{0.45cm} $\widehat{m_t} \gets m_t/(1-\beta_1^t)$, \hspace{0.2cm} $\widehat{v_t} \gets v_t/(1-\beta_2^t)$ \hspace{0.76cm} \textit{$\rhd$ Bias Correction}\\
    \hspace{0.5cm} $\theta_t \gets \theta_{t-1} - \alpha \widehat{m_t}/(\sqrt{\widehat{v_t}} + \epsilon)$ \hspace{2.17cm} \textit{ 	$\rhd$ Parameter Update}\\
    \textbf{return}  $\theta_{T}$
\label{algo:adabelief}
\end{algorithm}

\begin{algorithm}[!t]
\caption{Adabelief + Moment Centralization (AdabeliefMC) Optimizer}
\SetAlgoLined
\textbf{Initialize:} $\theta_{0},m_{0}\gets0,v_{0}\gets0,t\gets0$ \hspace{10mm} \textbf{Hyperparameters:} $\alpha, \beta_1, \beta_2, T$\\
\textbf{For} $\theta_{t}=$ [1,...,\textit{T}] \textbf{do}\\
    \hspace{0.45cm} $g_t \gets \nabla_{\theta} f_t(\theta_{t-1})$   \hspace{3.89cm} \textit{$\rhd$ Gradient Computation}\\
    \hspace{0.45cm} $m_t \gets \beta_1 m_{t-1} + (1-\beta_1) g_t$  \hspace{2.43cm} \textit{$\rhd$ First Order Moment}\\
    \hspace{0.45cm} \textcolor{blue} {$m_t \gets m_t  - mean(m_t)$}  \hspace{3.05cm} \textit{$\rhd$ \textcolor{blue}{Moment Centralization}}\\
    \hspace{0.45cm} $v_t \gets \beta_2 v_{t-1} + (1-\beta_2) (g_t-m_t)^2$ \hspace{1.49cm} \textit{$\rhd$ Second Order Moment} \\
    \hspace{0.45cm} $\widehat{m_t} \gets m_t/(1-\beta_1^t)$, \hspace{0.2cm} $\widehat{v_t} \gets v_t/(1-\beta_2^t)$ \hspace{0.8cm} \textit{$\rhd$ Bias Correction}\\
    \hspace{0.5cm} $\theta_t \gets \theta_{t-1} - \alpha \widehat{m_t}/(\sqrt{\widehat{v_t}} + \epsilon)$ \hspace{2.2cm} \textit{ 	$\rhd$ Parameter Update}\\
    \textbf{return}  $\theta_{T}$
\label{algo:adabeliefmc}
\end{algorithm}

\subsection{Datasets Used}
We test the performance of the proposed moment centralization optimizers on three benchmark datasets, including CIFAR10, CIFAR100 \cite{cifar} and TinyImageNet\footnote{http://cs231n.stanford.edu/tiny-imagenet-200.zip} \cite{le2015tiny}. The CIFAR10 dataset consists of 60,000 images from 10 object categories with 6,000 images per category. The 5,000 images per category are used for the training and the remaining 1,000 images per category are used for testing. Hence, the total number of images used for training and testing in CIFAR10 data is 50,000 and 10,000, respectively. The CIFAR100 contains the same 60,000 images of CIFAR10, but divides the images into 100 object categories which is beneficial to test the performance of the optimizers for fine-grained classification. The TinyImageNet dataset is part of the full ImageNet challange. It contsists of 200 object categories. Each class has 500 images for training. However, the test set consists of 10,000 images. The dimension of all the images is 64x64 in color in TinyImageNet dataset.

\subsection{Hyperparameter Settings}
All the optimizers in the experiment share the following settings. The decay rates of first and second order moments $\beta1$ and $\beta2$ are $0.9$ and $0.999$, respectively. 
The first and second order moments ($m$, $v$) are initialized to $0$. 
The training is performed for 100 epochs with $0.001$ learning rate for the first 80 epochs and $0.0001$ for the last 20 epochs.
The weight initialization is performed using random numbers from a standard normal distribution.

\begin{table}[!t]
    \caption{The results comparison of Adam \cite{adam}, Radam \cite{radam} and Adabelief \cite{adabelief} optimizers with the gradient centralization \cite{gc} and the proposed moment centralization on CIFAR10 dataset using VGG16 \cite{vgg} and ResNet18 \cite{resnet} CNN models. Note that the experiments are repeated three times with independent weight initialization.}
    \centering
    \begin{tabular}{|c|c|c|c|c|c|c|}
    \hline
    CNN Model & \multicolumn{3}{|c|}{VGG16 Model} & \multicolumn{3}{|c|}{ResNet18 Model}\\\hline\hline
    Optimizer & Adam & AdamGC & AdamMC & Adam & AdamGC & AdamMC \\\hline
    Run1 & 92.49 & 92.43 & 92.49 & 93.52 & 91.82 & 93.42\\
    Run2 & 92.52 & 90.64 & 92.26 & 93.49 & 93.43 & 93.37\\
    Run3 & 92.70 & 92.15 & 92.56 & 93.95 & 91.18 & 93.3\\\hline
    Mean$\pm$Std & \textbf{92.57}$\pm$0.11 & 91.74$\pm$0.96 & 92.44$\pm$0.16 & \textbf{93.65}$\pm$0.26 & 92.14$\pm$1.16 & 93.36$\pm$0.06\\\hline\hline
    
    Optimizer & Radam & RadamGC & RadamMC & Radam & RadamGC & RadamMC \\\hline
    Run1 & 92.79 & 92.16 & 93.33 & 94.02 & 93.67 & 94.00\\
    Run2 & 93.30 & 92.92 & 93.33 & 92.85 & 93.76 & 93.69\\
    Run3 & 92.80 & 92.58 & 93.36 & 94.06 & 93.7 & 94.08\\\hline
    Mean$\pm$Std & 92.96$\pm$0.29 & 92.55$\pm$0.38 & \textbf{93.34}$\pm$0.02 & 93.64$\pm$0.69 & 93.71$\pm$0.05 & \textbf{93.92}$\pm$0.21\\\hline\hline
    
    Optimizer & Adabelief & AdabeliefGC & AdabeliefMC & Adabelief & AdabeliefGC & AdabeliefMC \\\hline
    Run1 & 92.91 & 92.71 & 93.07 & 93.86 & 93.74 & 93.78\\
    Run2 & 92.67 & 93.04 & 92.94 & 92.31 & 93.88 & 93.72\\
    Run3 & 92.85 & 92.83 & 92.91 & 93.98 & 93.73 & 93.56\\\hline
    Mean$\pm$Std & 92.81$\pm$0.12 & 92.86$\pm$0.17 & \textbf{92.97}$\pm$0.09 & 93.38$\pm$0.93 & \textbf{93.78}$\pm$0.08 & 93.69$\pm$0.11\\\hline
    \end{tabular}
    \label{tab:results_cifar10}
\end{table}

\begin{table}[!t]
    \caption{The results comparison of Adam \cite{adam}, Radam \cite{radam} and Adabelief \cite{adabelief} optimizers with the gradient centralization \cite{gc} and the proposed moment centralization on CIFAR100 dataset using VGG16 \cite{vgg} and ResNet18 \cite{resnet} CNN models. Note that the experiments are repeated three times with independent weight initialization.}
    \centering
    \begin{tabular}{|c|c|c|c|c|c|c|}
    \hline
    CNN Model & \multicolumn{3}{|c|}{VGG16 Model} & \multicolumn{3}{|c|}{ResNet18 Model}\\\hline\hline
    Optimizer & Adam & AdamGC & AdamMC & Adam & AdamGC & AdamMC \\\hline
    Run1 & 67.74 & 67.68 & 68.47 & 71.4 & 73.74 & 74.48\\
    Run2 & 67.51 & 67.73 & 69.43 & 71.47 & 73.34 & 73.92\\
    Run3 & 67.96 & 68.26 & 68.01 & 71.64 & 73.58 & 74.64 \\\hline
    Mean$\pm$Std & 67.74$\pm$0.23 & 67.89$\pm$0.32 & \textbf{68.64}$\pm$0.72 & 71.5$\pm$0.12 & 73.55$\pm$0.2 & \textbf{74.35}$\pm$0.38\\\hline\hline
    
    Optimizer & Radam & RadamGC & RadamMC & Radam & RadamGC & RadamMC \\\hline
    Run1 & 69.56 & 70.31 & 70.89 & 73.54 & 73.75 & 74.31\\
    Run2 & 70.05 & 69.81 & 70.57 & 73.15 & 73.07 & 74.61\\
    Run3 & 69.82 & 70.26 & 70.35 & 73.41 & 73.88 & 74.48\\\hline
    Mean$\pm$Std & 69.81$\pm$0.25 & 70.13$\pm$0.28 & \textbf{70.60}$\pm$0.27 & 73.37$\pm$0.2 & 73.57$\pm$0.44 & \textbf{74.47}$\pm$0.15\\\hline\hline
    
    Optimizer & Adabelief & AdabeliefGC & AdabeliefMC & Adabelief & AdabeliefGC & AdabeliefMC \\\hline
    Run1 & 70.84 & 70.85 & 71.44 & 74.05 & 73.69 & 74.38\\
    Run2 & 70.71 & 71.17 & 70.83 & 74.11 & 73.95 & 74.86\\
    Run3 & 70.37 & 70.56 & 70.84 & 74.3 & 73.74 & 74.74\\\hline
    Mean$\pm$Std & 70.64$\pm$0.24 & 70.86$\pm$0.31 & \textbf{71.04}$\pm$0.35 & 74.15$\pm$0.13 & 73.79$\pm$0.14 & \textbf{74.66}$\pm$0.25\\\hline
    \end{tabular}
    \label{tab:results_cifar100}
\end{table}

\begin{table}[!t]
    \caption{The results comparison of Adam \cite{adam}, Radam \cite{radam} and Adabelief \cite{adabelief} optimizers with the gradient centralization \cite{gc} and the proposed moment centralization on TinyImageNet dataset using VGG16 \cite{vgg} and ResNet18 \cite{resnet} CNN models. Note that the experiments are repeated three times with independent weight initialization.}
    \centering
    \begin{tabular}{|c|c|c|c|c|c|c|}
    \hline
    CNN Model & \multicolumn{3}{|c|}{VGG16 Model} & \multicolumn{3}{|c|}{ResNet18 Model}\\\hline\hline
    Optimizer & Adam & AdamGC & AdamMC & Adam & AdamGC & AdamMC \\\hline
    Run1 & 42.38 & 45.02 & 43.50 & 49.08 & 53.90 & 53.02\\
    Run2 & 41.92 & 43.00 & 41.06 & 49.00 & 53.60 & 53.62\\
    Run3 & 42.72 & 44.42 & 39.22 & 49.00 & 52.54 & 54.10\\\hline
    Mean$\pm$Std & 42.34$\pm$0.4 & \textbf{44.15}$\pm$1.04 & 41.26$\pm$2.15 & 49.03$\pm$0.05 & 53.35$\pm$0.71 & \textbf{53.58}$\pm$0.54\\\hline\hline

    Optimizer & Radam & RadamGC & RadamMC & Radam & RadamGC & RadamMC \\\hline
    Run1 & 43.86 & 45.94 & 47.22 & 49.84 & 51.62 & 53.40\\
    Run2 & 44.10 & 45.86 & 47.48 & 50.50 & 51.54 & 52.72\\
    Run3 & 45.18 & 45.80 & 47.26 & 50.78 & 51.92 & 52.88\\\hline
    Mean$\pm$Std & 44.38$\pm$0.70 & 45.87$\pm$0.07 & \textbf{47.32}$\pm$0.14 & 50.37$\pm$0.48 & 51.69$\pm$0.20 & \textbf{53.00}$\pm$0.36\\\hline\hline
    
    Optimizer & Adabelief & AdabeliefGC & AdabeliefMC & Adabelief & AdabeliefGC & AdabeliefMC \\\hline
    Run1 & 47.32 & 47.84 & 47.22 & 51.82 & 52.02 & 53.04\\
    Run2 & 49.12 & 47.60 & 52.86 & 46.78 & 51.80 & 53.08\\
    Run3 & 47.16 & 47.94 & 48.34 & 51.04 & 51.56 & 53.78\\\hline
    Mean$\pm$Std & 47.87$\pm$1.09 & 47.79$\pm$0.17 & \textbf{49.47}$\pm$2.99 & 49.88$\pm$2.71 & 51.79$\pm$0.23 & \textbf{53.30}$\pm$0.42\\\hline
    \end{tabular}
    \label{tab:results_tinyimagenet}
\end{table}

\section{Experimental Results and Analysis}
\label{experimentalresults}
In order to demonstrate the improved performance of the optimizers with the proposed moment centralization, we conduct the image classification experiments using VGG16 \cite{vgg} and ResNet18 \cite{resnet} CNN models on CIFAR10, CIFAR100 and TinyImageNet datasets \cite{cifar} and report the classification accuracy. 
The results are compared with the corresponding optimization method without using moment centralization. Moreover, the results are also compared with the corresponding optimization method with the gradient centralization \cite{gc}. We repeat all the experiments three times with independent initializations and consider the mean and standard deviation for the comparison purpose. 
All results are evaluated with the same settings as described above. 

The classification accuracies on the CIFAR10, CIFAR100 and TinyImagenet datasets are reported in Table \ref{tab:results_cifar10}, Table \ref{tab:results_cifar100}, and Table \ref{tab:results_tinyimagenet}, respectively.
It is noticed from these results that the optimizers with moment centralization (i.e., AdamMC, RadamMC, and AdabeliefMC) outperform the corresponding optimizers without moment centralization (i.e., Adam, Radam, and Adabelief, respectively) and with gradient centralization (i.e., AdamGC, RadamGC, and AdabeliefGC, respectively) in most of the scenario. The performance of the proposed AdamMC optimizer is also comparable with Adam on the CIFAR10 dataset, where AdamGC (i.e., Adam with gradient centralization \cite{gc}) fails drastically.
It's also worth mentioning that the classification accuracy using the proposed optimizers is very consistent in different trials with a reasonable standard deviation in the results, except on TinyImageNet dataset using VGG16 model.


\begin{figure*}[t]
    \centering    
    \begin{subfigure}[t]{0.49\textwidth}
        \includegraphics[width=1\textwidth]{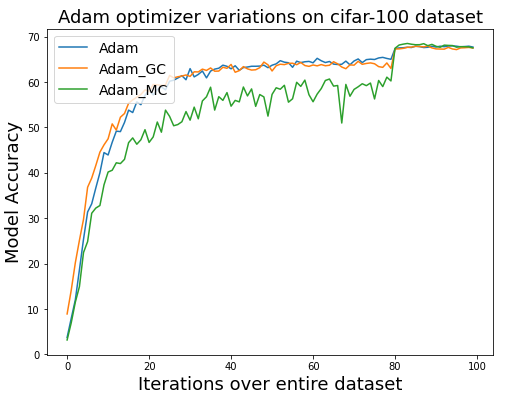}
        \caption{}
    \end{subfigure}
    \begin{subfigure}[t]{0.49\textwidth}
        \includegraphics[width=1\textwidth]{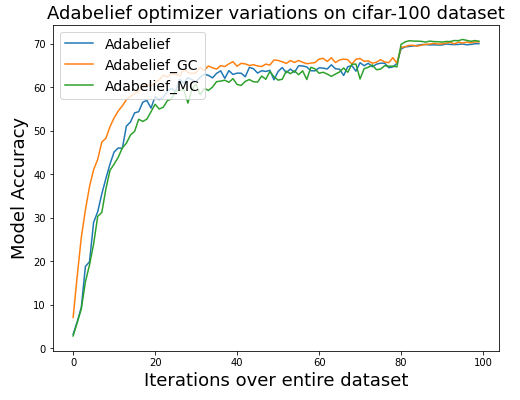}
        \caption{}
    \end{subfigure}
    \begin{subfigure}[t]{0.49\textwidth}
        \includegraphics[width=1\textwidth]{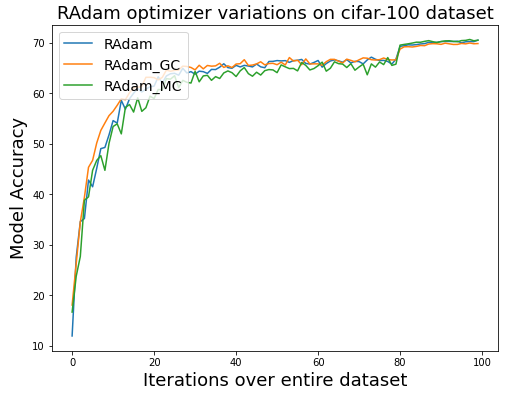}
        \caption{}
    \end{subfigure}
    \caption{The test accuracy vs epoch  plots using (a) Adam, Adam\_GC \& Adam\_MC, (b) Adabelief, Adabelief\_GC \& Adabelief\_MC, and (c) Radam, Radam\_GC \& Radam\_MC optimizers on CIFAR100 dataset.}
    \label{fig:accuracy}
\end{figure*}

The accuracy plot w.r.t. epochs is depicted in Fig. \ref{fig:accuracy} on CIFAR100 dataset for different optimizers.
Note that the learning rate is set to 0.001 for first 80 epochs and 0.0001 for last 20 epochs. It is noticed in all the plots that the performance of the proposed optimizers boosts significantly and outperforms other optimizers when the learning rate is dropped. It shows that the proposed moment centralization leads to better regularization and reaches closer to minimum.

\begin{figure*}[t]
\centering
\begin{subfigure}{0.49\textwidth}
    \includegraphics[width=1\textwidth]{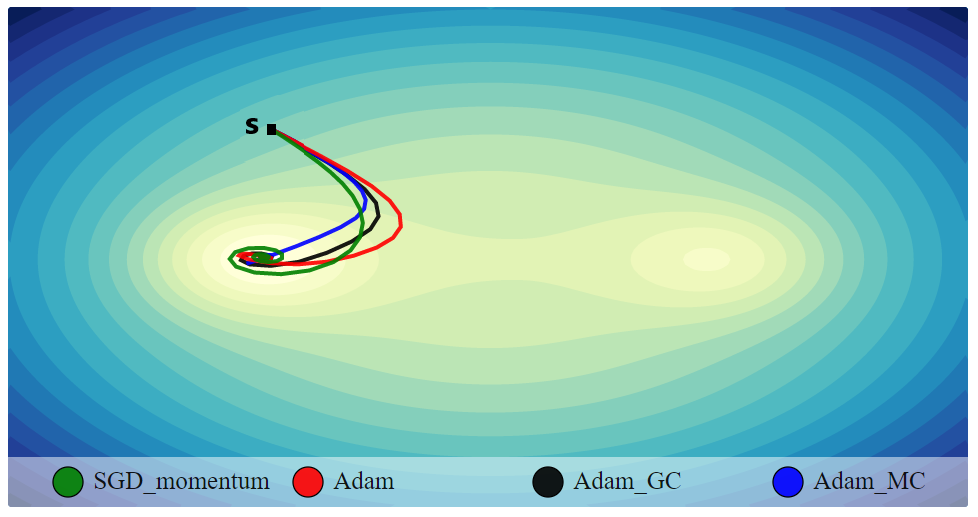}
    \caption{}
\end{subfigure}
\begin{subfigure}{0.49\textwidth}
    \includegraphics[width=1\textwidth]{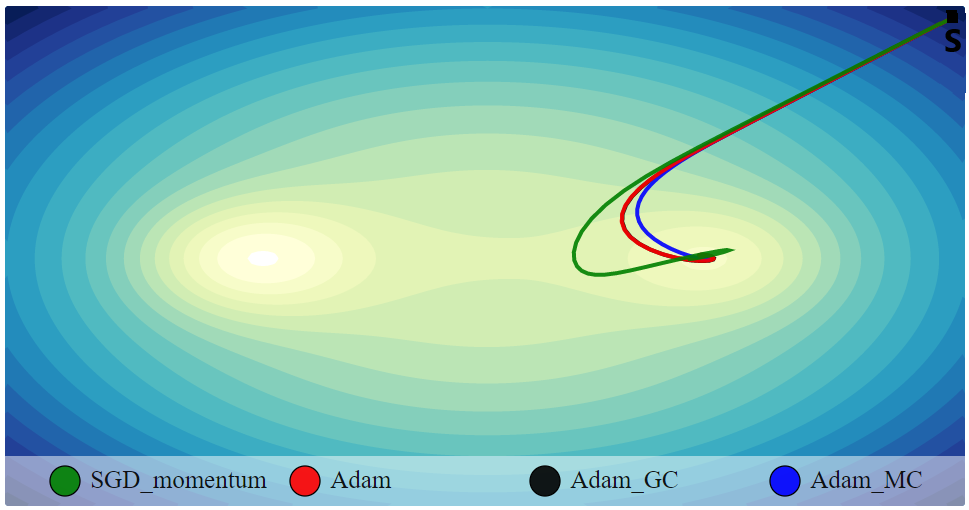}
    \caption{}
\end{subfigure}
\begin{subfigure}{0.49\textwidth}
    \includegraphics[width=1\textwidth]{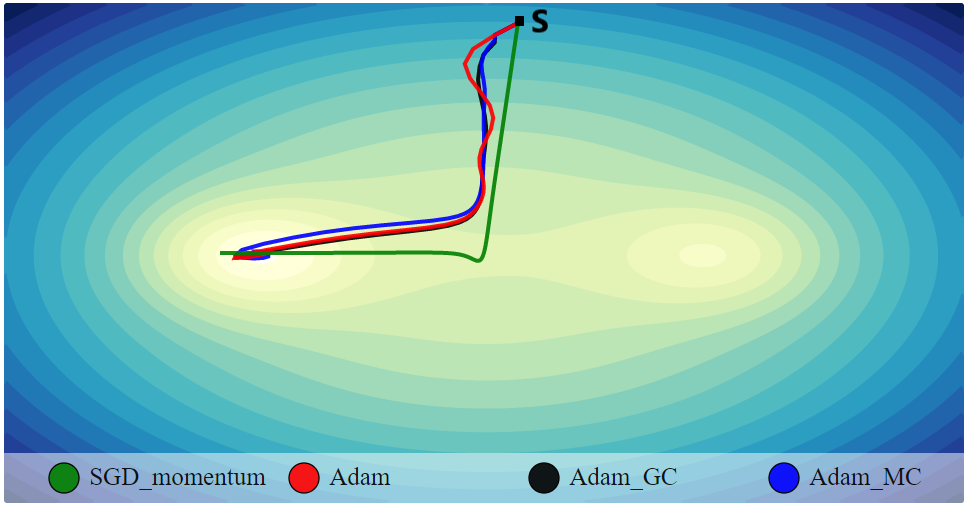}
    \caption{}
\end{subfigure}
\begin{subfigure}{0.49\textwidth}
    \includegraphics[width=1\textwidth]{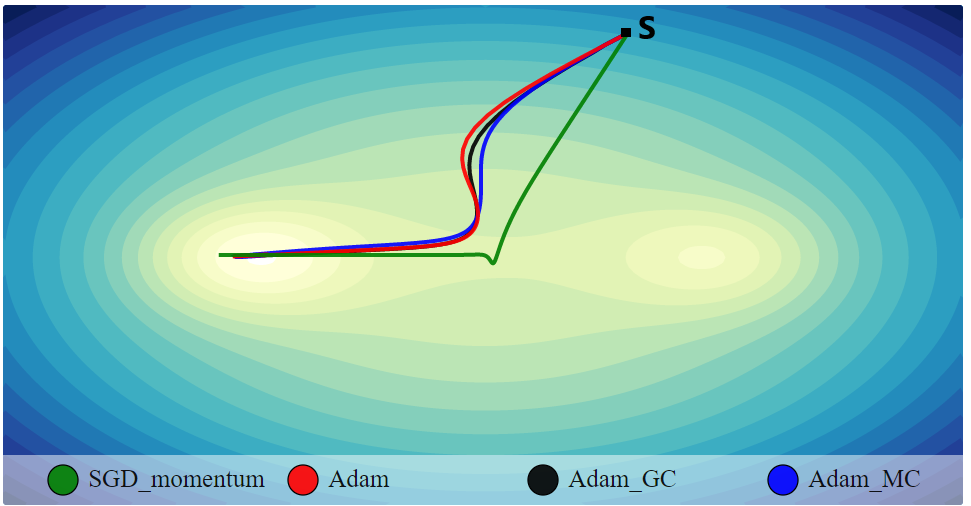}
    \caption{}
\end{subfigure}
\caption{The convergence of different optimization methods on a toy example with random initializations. \textbf{S} is the starting point.}
\label{fig:toy_example}
\end{figure*}

In order to justify the improved performance of the proposed Adam\_MC method, we show the convergence plot in terms of the optimization trajectory in Fig. \ref{fig:toy_example} with random intializations for following toy example:
\begin{equation}
f(x,y)=-2e^{\frac{-((x-1)^2 + y^2)}{2}} -3e^{\frac{-((x+1)^2 + y^2)}{2}} + x^2 + y^2.
\end{equation}
It is a quadratic `bowl' with two gaussian looking minima at (1, 0) and (-1, 0), respectively.
It can be observed in Fig. \ref{fig:toy_example}(a) that the Adam\_MC optimizer leads to a shorter path and faster convergence to reach minimum as compared to other optimizers. Figure 1 illustrates how the moment centralization for Adam, RAdam, and Adabelief achieved higher accuracy after the 80th epoch than without. This demonstrates that optimizers with MC variation have faster convergence speed than without. The SGD shows much oscillations near the local minimum. In Fig. \ref{fig:toy_example}(b), Adam and Adam\_GC depict fewer turns as compared to SGD with momentum, while Adam\_MC exhibits smoother updates. In Fig. \ref{fig:toy_example}(c-d), Adam\_MC leads to less oscillations throughout its course as compared to other optimizers. Hence, it is found that the proposed optimizer leads to shorter and smoother path as compared to other optimizers.


    
    

\section{Conclusion}
\label{conclusion}
In this paper, a moment centralization is proposed for the adaptive momentum based SGD optimizers. The proposed approach explicitly imposes the zero mean constraints on the first order moment. The moment centralization leads to better training of the CNN models. The efficacy of the proposed idea is tested with state-of-the-art optimizers, including Adam, Radam, and Adabelief on CIFAR10, CIFAR100 and TinyImageNet datasets using VGG16 and ResNet18 CNN models. It is found that the performance of the existing optimizers is improved when integrated with the proposed moment centralization in most of the cases. Moreover, the moment centralization outperforms the gradient centralization in most of the cases. Based on the findings from the results, it is concluded the moment centralization can be used very effectively to train the deep CNN models with improved performance. It is also observed that the proposed method leads to shorter and smoother optimization trajectory. The future work includes the exploration of the proposed idea on different types of computer vision applications.

\bibliographystyle{splncs04}
\bibliography{References}

\end{document}